# Unshredding of Shredded Documents: Computational Framework and Implementation

Lei Kristoffer R. Lactuan[1] and Jaderick P. Pabico[2]

*Institute of Computer Science, University of the Philippines Los Baños, College 4031, Laguna*

{[1]lkrlactuan, [2]jppabico}@up.edu.ph

**Abstract** – *A shredded document $D$ is a document whose pages have been cut into strips for the purpose of destroying private, confidential, or sensitive information $I$ contained in $D$. Shredding has become a standard means of government organizations, businesses, and private individuals to destroy archival records that have been officially classified for disposal. It can also be used to destroy documentary evidence of wrongdoings by entities who are trying to hide $I$ (e.g., as alleged by the whistle blowers in the P10-Billion pork barrel-JLN NGOs scam).*

*Shredding does not really destroy $I \in D$ but just simply jumbles $I$ by cutting the pages of $D$ into strips. Let $D$ be a set of $n$ ordered pages $\{p_0, p_1, \ldots, p_{n-1}\}$, and let each $i$th $(X \times Y)$-dimension page $p_{0 \leq i < n} \in D$ contains a 2-dimensional ordered arrangement of the information elements $J_{i,(x,y)} \in \{0,1\}$, $\forall\ 0 \leq x < X,\ 0 \leq y < Y$. When each $J_{i,(x,y)}$ is presented to a reader $R$ in correct order, then $I$ is transferred to $R$. When $p_i$ is shredded into $m$ vertical strips $s_{i,0}, s_{i,1}, \ldots, s_{i,m-1}$, then the $j$th strip $s_{i,0 \leq j < m}$ contains $X/m \times Y$ information elements, which has $J_{i,(a(j),0)}$ in its upper left corner and $J_{i,(b(j),Y-1)}$ in its lower right, where $a(r)=(x-m)r/m$, and $b(r) = (r+1)x/m$. However, what we are interested in are the information elements $J_{i,(a(j),0 \leq y < Y)}$ and $J_{i,(b(j),0 \leq y < Y)}$ along the left and right edges, respectively. If two strips $s_{i,p}$ and $s_{i,q}$ are actually adjacent to each other in the unshredded $p_i$, then either $J_{i,b(p),0 \leq y < Y} \equiv J_{i,(a(q),0 \leq y < Y)}$ or $J_{i,(a(p),0 \leq y < Y)} \equiv J_{i,(b(q),0 \leq y < Y)}$, where $\equiv$ is a similarity function that we defined. In this paper, we present an optimal $O((n \times m)^2)$ algorithm $A$ that reconstructs an $n$-page $D$, where each page $p$ is shredded into $m$ strips. We also present the efficacy of $A$ in reconstructing three document types: hand-written, machine typed-set, and images.*

I. INTRODUCTION

In the 2012 Corruption Perceptions Index (CPI), the Philippines ranked 105 out of 176 countries having a CPI score of 34/100, where a CPI score of 100 is perceived to be as corruption-free [1]. A rank of one is perceived to be the most corrupt, while a rank of 176 is perceived to be as the least corrupt country. The Philippines ranked fifth out of the 10 Southeast Asian countries on the same year. These 2012 rankings are lower compared to 2011 rankings in which the country ranked 129th internationally and seventh in Southeast Asia [2]. The country's corruption perception, and thus the country's CPI ranking, is expected to get worse in the election year of 2016 due to the surfacing of alleged corruption cases that triggered the resurfacing of other old corruption accusations. For example, the discovery of the 10-billion peso pork barrel scam triggered the resurfacing of the 2004 Fertilizer scam. It is alleged that the 10-billion peso pork barrel scam funneled public funds to several bogus non-governmental organizations (NGOs) under the name of one person in connivance with several high-ranking nationally-elected legislatures [3]. Recently, about 240 bank accounts and assets of the Philippine Vice President Jesus Jose "Jejomar" C. Binay, Sr. was ordered frozen by the Philippines' Court of Appeals on the petition of the Anti-Money Laundering Council from documentary evidences gathered by the Office of the Ombudsman [4].

Many fingers are being pointed at various people, as well as to various organizations, as either the mastermind or the beneficiaries of the alleged crimes. However, justice cannot be served to them because of lack of proper documentary evidences that will support the claims of the supposedly whistle blowers.

The whistle blowers allege that they themselves were once (knowingly or unknowingly) participants of the crime and could only narrate what had happened. This is because almost all, if not most, of these documentary evidences were destroyed to cover up the alleged crime [5]. Almost all of the oral narrations, however, corroborated each other which provided the investigating prosecutors the modes of operations of the alleged perpetuators and their cohorts [3].

Destroying documentary evidence may be done in several ways, such as by burning and by mechanical shredding. Documents that were burned, either with toxic chemicals, with water, or with fire, is very difficult if not impossible to reconstruct given the current state of technology that is available today. Shredded documents, however, may be reconstructed even if some of the shreds were already destroyed. For example, in 1979 when the United States Embassy in Iran was about to be taken over by Iranian students supporting the Iranian Revolution [6], the embassy personnel shredded their Iran-Contra documents. After the Embassy was taken over, the Iranians hired carpet weavers to reconstruct the shredded pieces of papers by hand [7].

Shredding an $n$-page document $D$ is the process of cutting each of the $n$ pages $p_{0 \leq i < n} \in D$ into $m$ rectangular pieces, called strips, by a mechanical shredder. The information $I$ contained in $D$ is supposedly destroyed because the pages have become unordered while the contents of each page were also unordered. However, given that all of the $n \times m$ strips are available, a very determined person can patiently reconstruct $D$ by hand, as in the example of the carpet weavers who were tasked to reconstruct shredded U.S. Embassy documents during the Iran-U.S. Hostage Crisis of 1979 [6]. The reconstruction may be intuitively done by taking any two strips $s_{i,p}$ and $s_{i,q}$ and visually match their edges to see if there is a continuity between them. Matching the edges of $s_{i,p}$ and $s_{i,q}$ takes four steps at the most (see Algorithm 1). The optimal reconstruction can be achieved via brute-force method that takes all possible pairs of $s_{i,p}$ and $s_{i,q}$ totaling to at most $(n \times m)^2$ combinations (or exactly $n^2 \times m^2 - n \times m$), resulting to at most $4 \times (n \times m)^2$ visual matches per pairwise combination of strips.

**Algorithm 1.** *Matching of two strips $s_{i,p}$ and $s_{i,q}$.*

| |
|---|
| 1) Match between the right edge of $s_{i,p}$ and the left edge of $s_{i,q}$ |
| 2) Match between the right edge of $s_{i,p}$ and the inverted right edge of $s_{i,q}$ |
| 3) Match between the inverted left edge of $s_{i,p}$ and the left edge of $s_{i,q}$ |
| 4) Match between the inverted left edge of $s_{i,p}$ and the inverted right edge of $s_{i,q}$ |

Although it has been reported that this problem has already been solved recently [8,9] through the announcement of the winner in Defense Advanced Research Projects Agency's (DARPA) Almost-Impossible Challenge [10], none of those who won, or even submitted solutions have published and made available to the public the details of their solutions [11-13]. We want to make public our detailed solution, and so in this paper, we present our computational framework for the automatic reconstruction of shredded documents. We alternatively used the term unshredding to mean the process of reconstruction. We then present our computational implementation which uses image processing techniques on images of strips as a pre-processing step. We then present our computational framework which we optimized to have a lesser number of steps than the intuitive solution mentioned above. We also present our results in unshredding various types of documents using our computational implementation. We tested the efficacy of our solution to reconstructing shredded papers of various types namely, hand-written, machine typed-set, and images.

II. COMPUTATIONAL FRAMEWORK

We discuss our framework under the assumption that we are given $n \times m$ unordered strips $s_{0 \leq i < n, 0 \leq j < m}$, where $n$ and $m$, as well as the indexes $i$ and $j$ are originally unknown. We can easily infer both $n$ and $m$ if we have prior knowledge of the dimension $X \times Y$ of the papers used and we are able to measure the length $x$ and height $y$ of each strip $s_{i,j}$. The strips have the trivial $y = Y$, while clearly, each page has $m = \lceil X/x \rceil$. To implement this framework, what we need to do now is (which we described in detail in the subsections that follow):

1. To scan all $n \times m$ strips $s_{0 \leq i < n, 0 \leq j < m}$ to convert them into a representation that the computer can automatically process;
2. To conduct image pre-processing to reduce the number of strips to consider, and therefore also reduce the number of pairwise combinations of strips;
3. To compute for the similarity **sim**($s_{i,p}$, $s_{i,q}$) of each pair and use this similarity metric to score a match between two strips as adjacent strips in the original unshredded $p_{0 \leq i < n}$;
4. To stitch all strips, using image processing techniques, according to the adjacency that was found by step three; and
5. To manually test the efficacy of the above steps under three document types: hand-written, machine typed-set, and images.

*A. Collecting and scanning of shredded papers*

We collected all strips of shredded papers and carefully laid them on a flatbed scanner, making sure that the strips do not touch or overlap each other as shown in Figure 1. To easily identify each strip for each scan step, we used a black background to delineate the strips in the image. We used a simple image processing technique [14,15] to identify the strips, cut each from the image, and save each to a file with $i$ and $j$ as identifiers. We will also refer to these scanned strips as $s_{0 \leq i < n, 0 \leq j < m}$ throughout this paper without loss of generality. For testing our framework, we simulated this step using the image of a page and manually shredded it with an aid of an image editor.

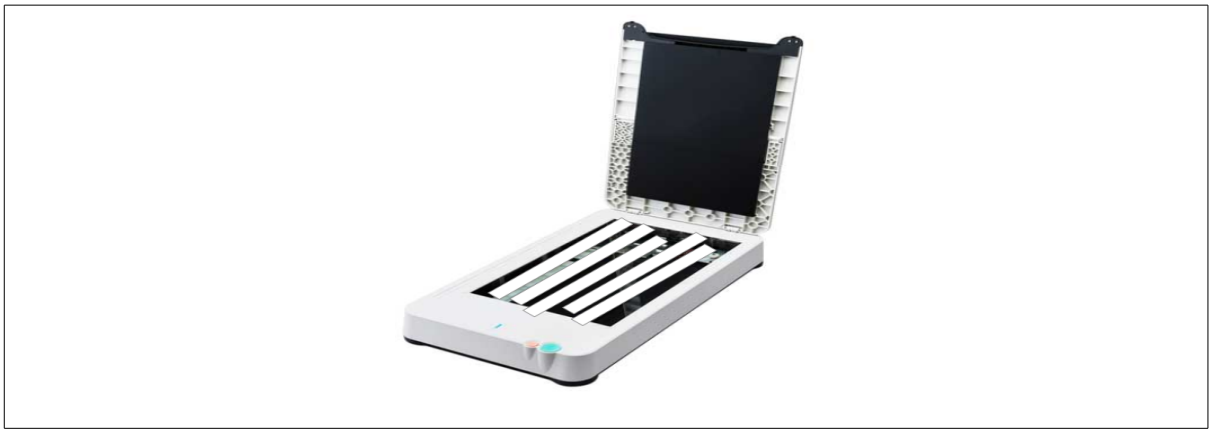

**Figure 1.** *How the shredded strips are supposed to be laid out onto the flatbed scanner. Notice the dark cover so that the background of the mostly white pages will stand out over the scanned strips.*

*B. Pre-processing of image strips*

To reduce the number of strips to evaluate, we used image processing to automatically remove all strips that will not be used in the reconstruction of pages. In this step, blank strips were removed because they will have no use in the process. Usually, these blank strips compose the margins of $p_{0 \leq i < n}$, unless, of course the margins themselves have annotations, doodles, or any identifying marks that will be useful in the reconstruction of the pages. We could have removed these blank strips in the scanning step described above but we included this step here because we were thinking of also automating that manual process.

We have also used printed character recognition techniques to set the strips into its upright position, whenever possible, which thus reduces the number of match steps described in Algorithm 1 down to 25%.

We then extracted the first two pixel columns at the left side of each strip, as well as that of the two pixel columns at the right side as four binary arrays. We termed them as $J_{i,(a(j),0 \leq y < Y)}$ and $J_{i,(a(j)+1,0 \leq y < Y)}$ for the left side, and as $J_{i,(b(j),0 \leq y < Y)}$ and $J_{i,(b(j)-1,0 \leq y < Y)}$ for the right side, where $a(r) = (x-m)r/m$, and $b(r) = (r+1)x/m$. We used these arrays to represent the edges of $s_{0 \leq i < n, 0 \leq j < m}$ in a data structure.

*C. Computation of similarity*

We defined a metric we called the similarity **sim**($s_{i,p}$, $s_{i,q}$) to score a match between two strips as adjacent strips in the original unshredded $p_{0 \leq i < n}$. We constructed a 4×4 array $Q$ composed of four adjacent rows of each of $J_{i,(a(j),0 \leq y < Y)}$, $J_{i,(a(j)+1,0 \leq y < Y)}$, $J_{i,(b(j),0 \leq y < Y)}$ and $J_{i,(b(j)-1,0 \leq y < Y)}$ such that, for example, $Q_{1,1} = J_{1,(b(j)-1,0 \leq y < Y)}$, $Q_{1,2} = J_{1,(b(j),0 \leq y < Y)}$, $Q_{1,3} = J_{1,(a(j),0 \leq y < Y)}$, and $Q_{1,4} = J_{1,(a(j)+1,0 \leq y < Y)}$. Using similarity of matrices, we matched $Q$ over each of the matrix template $T$, where some interesting ones are shown in Figure 2. Most other templates are either vertical or horizontal mirror images, horizontal or vertical translated images, or rotated versions of those that we have shown here.

Mathematically, $Q$ is similar to any of the $T$ if $Q = R^{-1}TR$, where $R$ is any 4×4 invertible similarity matrix. Of course, $R$ is invertible if $RR^{-1} = R^{-1}R = I_4$, where $I_4$ is the 4×4 identity matrix [16]. It is very easy to find $R$, and actually any $R$ can be used without affecting the outcome of the metric. The result of the metric, therefore, is the one that is in the neighborhood of $I_4$, if not exactly $I_4$. If **sim**($s_{i,p}$, $s_{i,q}$) $\equiv I_4$, then we stop computing for the similarities of all pairs that include the right side of $s_{i,p}$ or the left side of $s_{i,q}$, reducing further the number of pairs to be considered. We defined the symbol $\equiv$ to mean both sides are reduced to a scalar quantity that is 1 if $I_4$, otherwise >1 if just in the neighborhood of $I_4$.

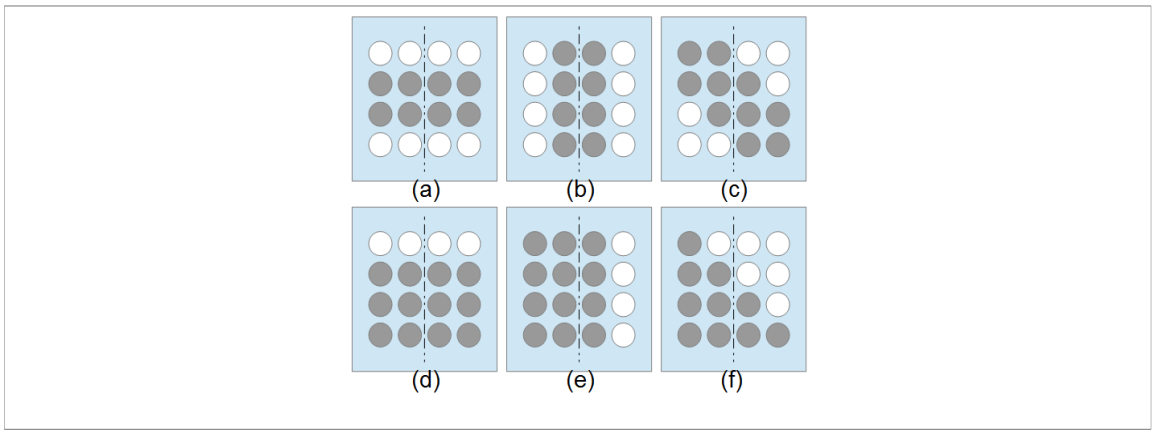

**Figure 2**. Some interesting templates that we used in the similarity test. Shown are templates for (a) a horizontal line, (b) a vertical line, (c) a diagonal line, (d) a horizontal edge of a polygon part, (e) a vertical edge of a polygon, and (f) a diagonal edge of a polygon part. Dark colored circles have a value of 1 in the binary matrix, while light colored ones have 0.

*D. Stitching of matched strips into pages*

We considered two strips $s_{i,p}$ and $s_{i,q}$ with the highest similarity **sim**($s_{i,p}$, $s_{i,q}$) as adjacent strips. Using image processing techniques [17,18], we stitched $s_{i,p}$ and $s_{i,q}$ together in that order. We considered the stitching of any page $p_{0 \leq i < n}$ completed if we have already stitched at most *m* strips together.

*E. Manual Evaluation*

We evaluated our method by reconstructing three different types of documents: hand-written, machine typed-set, and images. Since our methodology is heavily biased on the machine typed-set (e.g., printed character recognition technique described in II.B), we hypothesized that this type of document will be reconstructed with less error. We only used a manual method for evaluating the reconstructed documents since we lack a metric for automatically doing this.

III. EVALUATION RESULTS

Figure 3a shows the images of sample documents before they were shredded. These images are examples from among the many that we evaluated. Figure 3b shows the images of the stitched documents after the documents in Figure 3a have been shredded and undergone our computational framework. Since we have removed the blank strips, which mostly compose the margins of the original documents, it can be clearly seen that the reconstructed pages lack margins. Margins, however, are not important since what we want to reconstruct are the information that are contained in the pages, unless, of course, the margins themselves contain information, such as, for example editor's annotations, doodles, or other identifying marks. The extra information in the margins would have been caught by our framework.

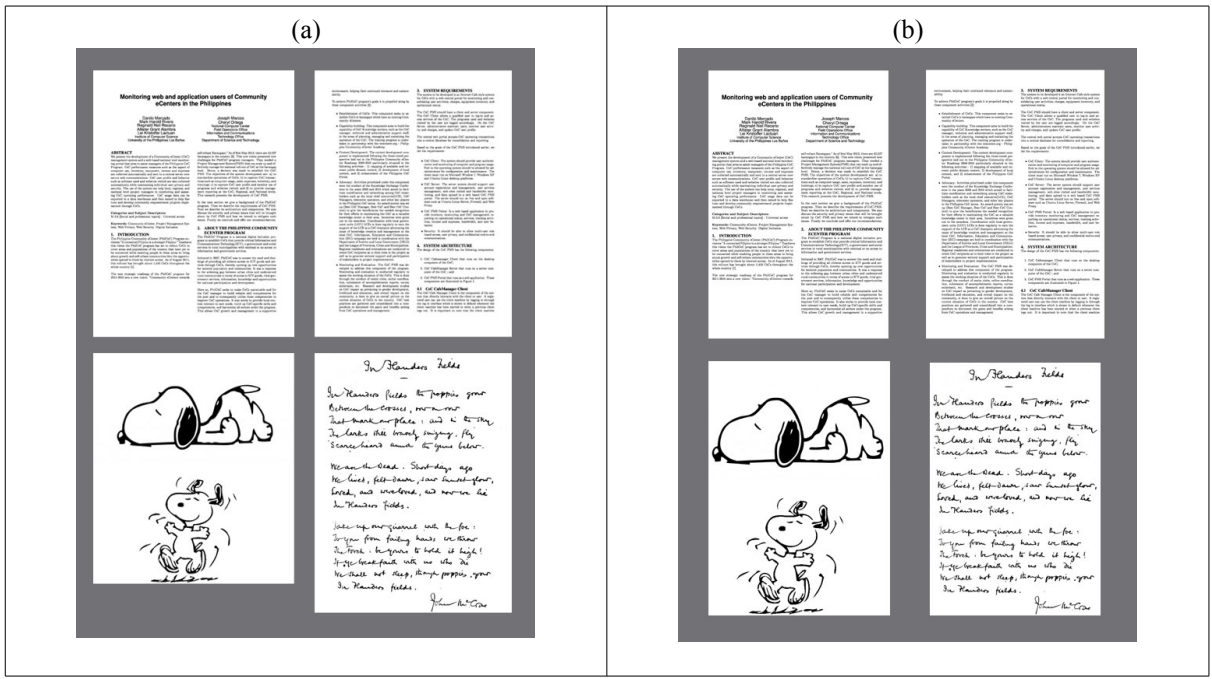

**Figure 3.** Sample documents of each type (a) before shredding, and (b) after reconstructing using our computational framework.

We tested our framework using a subjective evaluation of the human users. Although there exist several advanced optical character and image recognizers (see for example [19]), as well as state-of-the-art natural language processors (see for example [20]), we did not automatically, and thus, objectively evaluate the output of our framework using these advanced systems because of their respective inherent implementation complexities [21]. Intuitively, the subjective evaluations would have sufficed to help people automatically unshred the shredded documents.

IV. SUMMARY AND CONCLUSION

In this paper, we presented the computational framework and implementation of an automated process for reconstructing shredded documents. We initially estimated that the process will take $4 \times (n \times m)^2$ matches, which we reduced to about $(n \times m)^2$ due to our implementation of the optical character reader. We further reduced n and m by expunging from the set of strips those that are blanks. These blank strips are usually parts of the paper margins. We subjectively evaluated the results of our framework and we found out that it is able to unshred shredded papers by stitching two strips $s_{i,p}$ and $s_{i,q}$ together if their respective edges have high similarity scores **sim**($s_{i,p}$, $s_{i,q}$). We defined our similarity metric using the matrix similarity principles. We further reduced the number of matches in our framework if the two strips si,p and si,q score a similarity that matches the identity matrix $I_4$.

V. ACKNOWLEDGMENTS, DISCLOSURE OF EARLIER PRESENTATION, AND AUTHOR CONTRIBUTIONS

This research effort is funded by the Institute of Computer Science (ICS) Core Fund conducted during the First and Second Semesters of AY 2013-2014 under the directorship of *Prof. Jaime M. Samaniego*. We used the Application Programmer Interface in the ImageLab Library of Image Processing Functions provided for academic and research purposes by *Prof. Vladimir Y. Mariano, Ph.D.*, Associate Professor at ICS.

An earlier version of this paper was presented as an oral paper during the *6th Annual University of the Philippines Los Baños (UPLB) College of Arts and Sciences (CAS) Student-Faculty Research Conference (SFRC 2013)* held at the NCAS Auditorium, UPLB on 13 December 2013.

The following are the respective contributions of the authors: (1) JPP formulated the computational solution to the problem; (2) LKRL implemented the computational solution; (3) Both LKRL and JPP conducted the computational experiments and the performance analyses; and (4) Both LKRL and JPP prepared and edited the final manuscript. Both authors declare no conflict of interest.